\documentclass[sigconf]{acmart}

\AtBeginDocument{%
  \providecommand\BibTeX{{%
    \normalfont B\kern-0.5em{\scshape i\kern-0.25em b}\kern-0.8em\TeX}}}
\setcopyright{acmcopyright}
\copyrightyear{2019}
\acmYear{2019}
\acmDOI{}

\acmConference[KDD '19]{KDD '19: 25th ACM SIGKDD Conference on Knowledge Discovery and Data Mining}{August 05, 2019}{Anchorage, AK}
\acmBooktitle{KDD '19: 25th ACM SIGKDD Conference on Knowledge Discovery and Data Mining,
  August 05, 2019, Anchorage, AK}
\acmPrice{}
\acmISBN{}

\begin{document}

\title{Improving ML Training Data with Gold-Standard Quality Metrics}

\author{Leslie Barrett}

\affiliation{%
  \institution{Bloomberg LP}
  \country{USA}}
\email{lbarrett4@bloomberg.net}

\author{Michael W. Sherman}
\affiliation{%
  \institution{Google}
  \country{USA}}
\email{michaelsherman@google.com}








\begin{abstract}
Hand-tagged training data is essential to many machine learning tasks. However, training data quality control has received little attention in the literature, despite data quality varying considerably with the tagging exercise. We propose methods to evaluate and enhance the quality of hand-tagged training data using statistical approaches to measure tagging consistency and agreement. We show that agreement metrics give more reliable results if recorded over multiple iterations of tagging, where declining variance in such recordings is an indicator of increasing data quality. We also show one way a tagging project can collect high-quality training data without requiring multiple tags for every work item, and that a tagger burn-in period may not be sufficient for minimizing tagger errors.
\end{abstract}




\maketitle

\section{Background and Problem}
Human-tagged\footnote{Many terms are commonly used to refer to a human manually enriching data, including "annotator", "rater", "coder", "tagger" and "labeler". For consistency we arbitrarily use "tagger" and related forms, other than when referring to "inter-rater agreement".} training data is essential to supervised machine learning. In the last few years, services like Mechanical Turk and Figure Eight (formerly known as CrowdFlower) have increased the ease of  collecting hand-tagged data, but assessing the quality of tags remains an issue with taggers of all levels of expertise. With text in particular, ambiguity in the underlying data or the tagging instructions may cause tagging inconsistencies that affect the performance of machine learning models trained on the tagged data. 

This problem of inconsistent tagged data has been addressed in the research community, resulting in advances in statistical measures of consistency and agreement among human taggers. Following Carletta \cite{carletta1996assessing}, inter-rater agreement on language-tagging tasks has taken distributional data effects into account rather than simply comparing the percentage of overlapping tags. In particular, Cohen's kappa \cite{cohen1960coefficient} was introduced as an agreement metric for tagging natural language data.

Further research revealed deficiencies in Cohen's kappa due to its inability to account for disagreements in the data despite accounting for distributional effects. Kappa is prone to inconsistencies in the presence of skewed data, giving rise to the "paradox problem" \cite{feinstein1990high}, and does not handle missing values. Krippendorff \cite{krippendorff1980content} addressed these problems with the introduction of Krippendorff's alpha, a metric incorporating disagreement among taggers which allows for for missing data and multi-tagger scenarios. In particular Antoine et al. \cite{antoine2014weighted} showed that Krippendorff's alpha \cite{krippendorff1980content} is a more reliable metric than kappa on natural language tagging tasks including emotions, opinions and co-references. Krippendorff's alpha (Figure \ref{fig:kripp}) has since been widely adopted in the NLP community and has become the standard for tagging tasks based on language data. 
\begin{figure}[h]
\caption{Krippendorff's Alpha}
\centering
\begin{displaymath}
A = \frac{D_e-D_o}{D_o}
\end{displaymath}
Where $D_o$ is the observed disagreement between taggers and $D_e$ is the expected or chance disagreement.
\Description{Krippendorff's alpha equals the total of expected disagreement - the observed disagreement between taggers, divided by the observed disagreement.}
\label{fig:kripp}
\end{figure}

\begin{figure*}[h]
\caption{Tagging Task Design Process}
\centering
\includegraphics[width=\textwidth]{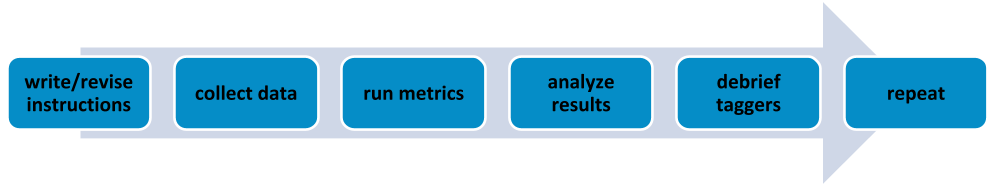}
\Description{Shows the steps of the tagging task design process: write/revise instructions, collect data, run metrics, analyze results, debrief taggers, repeat}
\label{fig:process}
\end{figure*}

Many current studies take a single measurement of inter-rater agreement to validate a training set or provide an upper-bound on model accuracy. Generally, the main consideration is that the resulting tags are "reliable" and reproducible \cite{artstein2008inter}. Others \cite{schaer2012better} use the results to alter the data, removing samples that are problematic for taggers. 

Few studies have focused on determining the degree of underlying ambiguity in the data itself. Dumitrace et al.\cite{dumitrache2018crowdtruth} create a model to represent a crowdsourcing system in three main components---workers, input units and tags. Their model proposes to explain how noise in any one of the three components influences the other components and the overall tagged data quality. In particular, it is one of the only studies to separate the noise generated by taggers from noise due to the inherent ambiguity of the data and the tagging task itself.

In the present study, we build upon this idea by attempting to reduce noise due to the tagging task and the taggers as much as possible, leaving only noise due to data ambiguity. Additionally, we specifically avoid replacing taggers or tagging every work item multiple times. These two common techniques for increasing tag quality suffer from the assumptions that tag creation is low cost, and that tagging task administrators are overly willing to replace individual taggers. Neither of these assumptions is necessarily the case, especially when a tagging task requires special expertise.

To create high-quality tagged data with these restrictions, we use observations of inter-rater agreement in three different contexts throughout the data collection process. First, we use inter-rater agreement measurements to design the tagging task. Next, we use inter-rater agreement measurements to determine when our taggers are fully educated\footnote{We use the term "education" and related forms to describe tagger instruction rather than "training", to avoid confusion with "training data".} on the tagging task (rather than a fixed-length burn in period). Finally, we use inter-rater agreement measurements  to monitor our educated taggers through the collection of the tagged data. Taking inter-rater agreement measurements in these three contexts allow for the creation of high quality data when practical concerns make collecting multiple tags on the entire dataset infeasible, while also reducing the risk associated with taking only a single inter-rater agreement measurement.

\section{Methodology and Data}

\subsection{Tagging Task Description}
In our tagging task, taggers were given pairs of sentences and asked to rate the quality of the paraphrase between the two sentences on a 1-4 scale, where 1 represents no similarity and 4 represents a perfect paraphrasing. This use of an ordinal tag to quantify shared meaning between two sentences is described in previous literature as Semantic Textual Similarity \cite{agirre2012semeval} and is closely related to paraphrase detection. 

Our untagged data is from a corpus of legal text. This resulted in two specific challenges. First, we could not rely on general notions of "paraphrase" and "similarity", as these have different meanings in the legal domain than in a general context. Second, legal expertise was required to meaningfully tag the data, so taggers needed to be sourced independently (at considerable expense) rather than relying on crowdsourcing platforms.

\subsection{Tagging Task Design}

Before we began collecting large amounts of data, we wanted to make sure our tagging task was well specified enough that multiple taggers would give most sentence pairs identical scores. To accomplish this, we went through five rounds of tagging task design (Figure \ref{fig:process}). Each round started with a set of instructions for the tagging task, followed by two experienced attorneys using the instructions to tag a set of data. We then analyzed the tagged data (including inter-rater agreement metrics). The analysis was used to debrief the attorneys, with a focus on determining what instruction misunderstandings led to disagreed tags. Finally, the insights from the debrief were used to create a new set of instructions for the next round.

After achieving a Krippendorff's alpha above .8 (.889) on the fifth round, followed by a debrief with positive feedback from the attorneys, we decided our tagging task was clear enough to proceed. Krippendorff's >.8 is considered "almost perfect agreement" based on the Landis \& Koch scale \cite{landis1977measurement}, a commonly used benchmark for interpretation.

\subsection{Tagger Education}

\begin{figure}[h]
\caption{Krippendorff's Alpha of Tagger Education Data}
\centering
\includegraphics[width=\columnwidth]{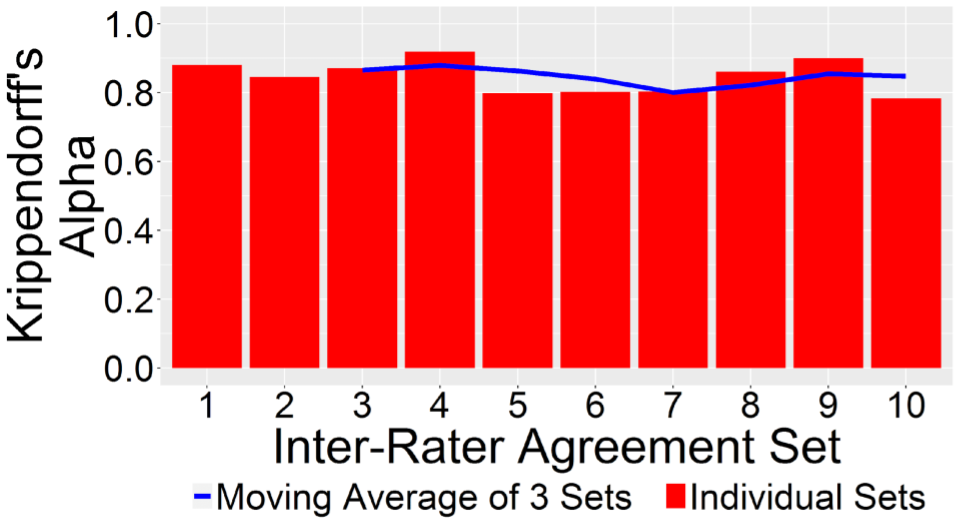}
\Description{Shows a bar chart with 10 Krippendorff's alpha scores all around .8, with a moving average line also around .8 starting with the 3rd bar.}
\label{fig:fig3}
\end{figure}

For the tagging task design we used experienced attorneys as taggers, but this was financially infeasible for collecting a large amount of data. Instead, a group of five less experienced legal services professionals was engaged for subsequent tagging. First, this group was educated by the experienced attorneys who had done the tagging for the task design. Next, we recorded Krippendorff's alpha on 10 sets of "education data" to ensure the group of five taggers was producing data of sufficient quality. Each of these 10 sets of tagger education data had 50 sentence pairs (for a total of 500 sentence pairs), with every pair tagged by every tagger. While there were inconsistencies in Krippendorff's alpha across the education data sets, Figure \ref{fig:fig3} shows that a 3-set moving average of Krippendorff's alpha was always above .8, suggesting the group of five taggers was reliability tagging data. This "burn-in" phase has become common practice with crowdsourced tagging \cite{sabou2014corpus}. 

\subsection{Data Collection and Tagger Monitoring}

The tagging of the main dataset (35136 sentence pairs) took place over three months. Every sentence pair in the main dataset was tagged by only a single tagger. To ensure the quality of the data remained high, we collected smaller "monitoring datasets" (ranging from 60 to 150 sentence pairs) twice each week. Each sentence pair in a monitoring dataset was tagged by multiple taggers. The number of sentence pairs in the monitoring datasets assigned to each tagger was adjusted to correspond to the proportions of the main dataset tagged by each tagger, since the taggers worked at different speeds. Discrepancies in the monitoring data were inspected by experienced attorneys, who then instructed individual taggers on how they could improve.

\begin{figure}[h]
\caption{Krippendorff's Alpha of Monitoring Datasets}
\centering
\includegraphics[width=\columnwidth]{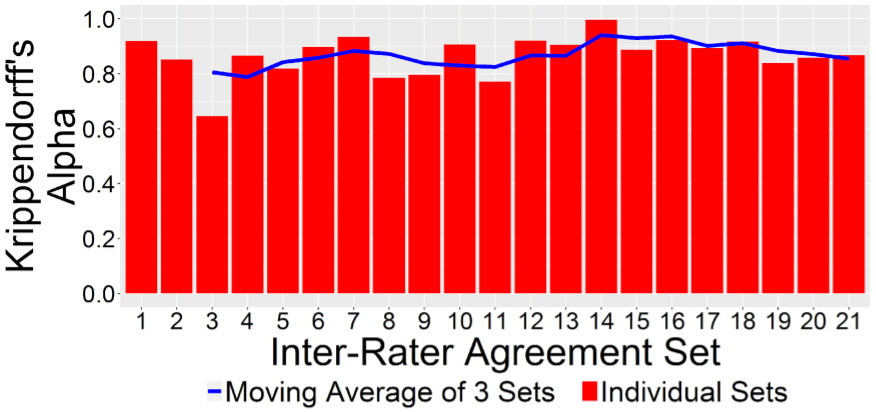}
\Description{Shows a bar chart with 21 Krippendorff's alpha scores all around .8 but with considerable variance, with a moving average line also around .8 starting with the 3rd bar.}
\label{fig:fig4}
\end{figure}

\begin{figure}[h]
\caption{Moving Variance of Krippendorff's Alpha of Five Monitoring Datasets}
\centering
\includegraphics[width=\columnwidth]{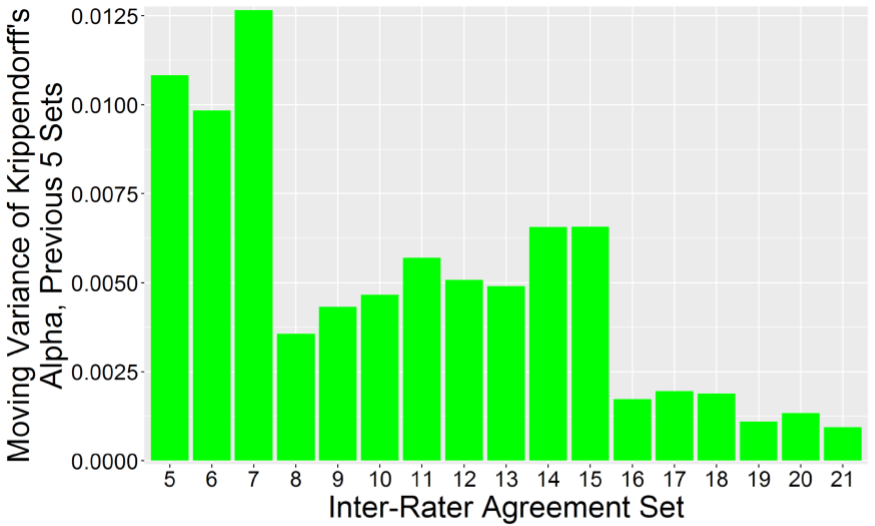}
\Description{Shows a bar chart with 21 variances, starting with set 5. The first 3 variances are around .01, the next 8 around .005, and the final  around .0015.}
\label{fig:fig5}
\end{figure}

Krippendorff's alpha scores for the 21 monitoring datasets and a moving average are in Figure \ref{fig:fig4}. We also calculated the moving variance of the Krippendorf's alpha values from the five previous monitoring datasets, which is seen in Figure \ref{fig:fig5}.

\section{Discussion}

Figure \ref{fig:fig3} shows relatively consistent performance by all five taggers during their education. This suggests the time invested in tagger education and tagging task design was well spent, as there is no obvious gain in agreement across these 500 tagged sentence pairs of education data. This led us to believe burn-in was complete and the tagging task was fully understood by the taggers. Initially, the alpha scores on the monitoring datasets in Figure \ref{fig:fig4} appeared to confirm this.

However, after examining the moving variances in Figure \ref{fig:fig5}, the taggers continue to improve throughout the collection of the main dataset. Variance drops at both the 8th and 16th monitoring dataset, and appears to converge at a very small value at the 20th monitoring dataset, despite mostly high (>.8) values of Krippendorff's alpha for most of the monitoring datasets (Figure 4). This tells us that burn-in was not fully complete until very far into the tagging process.

We believe the primary driver of decreasing moving variance of Krippendorff's alpha in Figure \ref{fig:fig5} was the review of disagreeing tags in the monitoring datasets by experienced attorneys. Each example of disagreement in the monitoring datasets was considered by an experienced attorney, who then provided feedback to taggers about how they could improve. Although none of these interventions were dramatic enough to result in changes to the tagging task, these interventions did result in more consistent tagging by the five taggers based on the reduced moving variance of Krippendorff's alpha as seen in Figure \ref{fig:fig5}. We hypothesize that specific interventions around the 3rd and 11th monitoring datasets were especially useful to the taggers, although we have no record of the content of conversations between the experienced attorneys and taggers.

The data further show the importance of sampling inter-rater agreement more than once when collecting multiple tags on every work item is infeasible. In this tagging task, not only was the variance of Krippendorff's alpha across multiple monitoring datasets initially high, but agreement improved over time as variance declined.

Our results also suggest there is a minimum threshold for agreement on a given dataset. This suggests common interpretations of "strength of agreement", like the one proposed in Landis and Koch \cite{landis1977measurement}, may only be appropriate in limited contexts where the data and tagging task are unambiguous.

We note that some of our sample sizes were small, potentially biasing Krippendorff's alpha towards outliers. In other words, drawing too few sentence pairs for each monitoring dataset could result in a monitoring dataset with a balance of "easy" and "difficult" sentence pairs different than the balance in the entire data. In the future we would consider collecting monitoring datasets with a larger number of work items, although we believe using moving metrics of agreement statistics (rather than considering each monitoring dataset's agreement individually) mitigates some of the impact of outliers.

\section{Conclusion}

Our results show it is possible to improve the quality of a human-tagged paraphrase detection dataset through multiple rounds of inter-rater agreement analysis with tagger-specific interventions based on disagreed work items. Our results also show the rolling variance of Krippendorff's alpha on monitoring datasets decreased as the taggers became more experienced. As this variance decreased, the agreement values came closer to reflecting the "true" ambiguity in the data as opposed to ambiguity contributed by outside factors like poor instructions and inadequate tagger education.

The continued drop in moving variance of Krippendorff's alpha across multiple monitoring datasets implies conventional approaches to burn-in, which focus primarily on tagging a small number of work items at the beginning of a data collection project, may not be sufficient when dealing with highly ambiguous data and/or a difficult tagging task (both which are common with language data). Additionally, our results suggest that tagger education plus a burn-in period may not maximize tagger performance without continued monitoring, and that continued monitoring of taggers reveals tagger-improving interventions that an initial analysis could miss.

Furthermore, we show a feasible alternative to having multiple tags on each work item by conducting periodic agreement studies as described. The collection of tagged data is often a cost barrier to building machine learning models, and our results show a considerable amount of savings is possible without compromising data quality.

In the future we would repeat this experiment for other tagging tasks with a view to eventually developing a more robust interpretation scale for agreement metrics like Krippendorff's alpha that take into account ambiguities in the data. For example a "reliable" result may be one in which the metric's variance is reduced by a certain amount over a given set of iterations rather than a fixed point that applies unilaterally.

\bibliographystyle{ACM-Reference-Format}
\bibliography{bibliography}

\end{document}